# Debiasing pipeline improves deep learning model generalization for X-ray based lung nodule detection

**PREPRINT**

**Michael J. Horry [a,b], Subrata Chakraborty [a,c,\*], Biswajeet Pradhan [a,d,e,\*], Manoranjan Paul [f], Jing Zhu [g], Hui Wen Loh [h], Prabal Datta Barua [a,i,j], U. Rajendra Arharya [h,j,k,l,m]**

[a] Faculty of Engineering and Information Technology, University of Technology Sydney, Ultimo, NSW 2007, Australia

[b] IBM Australia Limited, Sydney, NSW 2000, Australia

[c] Faculty of Science, Agriculture, Business and Law, University of New England, Armidale, NSW 2351, Australia

[d] Center of Excellence for Climate Change Research, King Abdulaziz University, Jeddah 21589, Saudi Arabia

[e] Earth Observation Center, Institute of Climate Change, Universiti Kebangsaan Malaysia, Selangor 43600, Bangi, Malaysia

[f] Machine Vision and Digital Health (MaViDH), School of Computing and Mathematics, Charles Sturt University, Bathurst, NSW 2795, Australia

[g] Department of Radiology, Westmead Hospital, Westmead, NSW 2145, Australia

[h] School of Science and Technology, Singapore University of Social Sciences, Singapore 599494, Singapore

[i] Cogninet Brain Team, Cogninet Australia, Sydney, NSW 2010, Australia

[j] School of Business (Information Systems), Faculty of Business, Education, Law & Arts, University of Southern Queensland, Toowoomba, QLD 4350, Australia

[k] School of Engineering, Ngee Ann Polytechnic, Singapore 599489, Singapore

[l] Department of Bioinformatics and Medical Engineering, Asia University, Taichung 413, Taiwan

m Research Organization for Advanced Science and Technology (IROAST), Kumamoto University, Kumamoto 860-8555, Japan

\*       Correspondence: Subrata.Chakraborty@uts.edu.au; Biswajeet.Pradhan@uts.edu.au

**Abstract**

*Background:* Lung cancer is the leading cause of cancer death worldwide by a large margin and a good prognosis can only be provided when the cancer is diagnosed early. Unfortunately, screening programs for the early diagnosis of lung cancer are uncommon. This is in-part due to the at-risk groups being located in rural areas far from medical facilities. Reaching these populations would require a scaled approach that combines mobility, low cost, speed, accuracy, and privacy. We can resolve these issues by combining the chest X-ray imaging mode with a federated deep-learning approach, provided that the federated model is trained on homogenous data to ensure that no single data source can adversely bias the model at any point in time. In this study we show that an image pre-processing pipeline that homogenizes and debiases chest X-ray images can improve both internal classification and external generalization, paving the way for a low-cost and accessible deep learning-based clinical system for lung cancer screening.

*Methods:* An evolutionary pruning mechanism is used to train a nodule detection deep learning model on the most informative images from a publicly available lung nodule X-ray dataset. Histogram equalization is used to remove systematic differences in image brightness and contrast. Model training is performed using all combinations of lung field segmentation, close cropping, and rib suppression operators. We show that this pre-processing pipeline results in deep learning models that successfully generalize an independent lung nodule dataset using ablation studies to assess the contribution of each operator in this pipeline.

*Conclusion:* In stripping chest X-ray images of known confounding variables by lung field segmentation, along with suppression of signal noise from the bone structure we can train a highly accurate deep learning lung nodule detection algorithm with outstanding generalization accuracy of 89% to nodule samples in unseen data.

## 1 Introduction

Lung cancer is the leading cause of cancer death worldwide with 1.80 million deaths documented by the World Health Organization in 2020 (WHO, 2021a). The global deaths worldwide attributable to lung cancer are twice that of the second most common cause of cancer death, colorectal cancer. Studies have shown a favorable prognosis for early-stage lung cancer with 5-year survival rates of up to 70% for patients with small, localized tumors (Blandin Knight et al., 2017). Relieving the economic and sociological costs associated with lung cancer is therefore dependent on the early diagnosis of this condition.

A comprehensive meta-review conducted by (Sadate et al., 2020) concluded that low-dose CT (LDCT) lung cancer screening can reduce lung cancer associated mortality by 17% and overall mortality by 4% in risk populations that are highly exposed to tobacco. Although LDCT is more effective at detecting lung cancer at an early stage than Chest X-ray (CXR) (Henschke et al., 1999), the feasibility and cost-effectiveness of LDCT screening in low and middle-income countries without well-developed healthcare infrastructure has been called into question (Edelman Saul et al., 2020). In some rural areas lack of transportation, long distances and poor road conditions can make healthcare inaccessible for many (Bray et al., 2018). Compounding this problem is the relative lack of skilled radiologists and oncology professionals per head of population, particularly in low and middle-income countries. Even in high-income countries, LDCT is known for its tendency to produce false positives, resulting in invasive procedures aimed at characterizing the nodule and associated false positive nodule workup risks (Dajac et al., 2016). For these reasons, lung cancer screening remains uncommon with only two countries currently implementing lung cancer screening being the USA and China. These programs target high-risk lifetime smokers, in contrast to the entire population screening programs that are common for other cancers such as breast, cervical and colorectal (Pinsky, 2018).

Extending the scope of lung cancer screening to cover a broader population depends on increasing community access to screening facilities at affordable cost. CXR is a widely available, safe, simple to operate, and inexpensive medical imaging technology in comparison to LDCT (Shankar et al., 2019). CXR image acquisition apparatus is readily available in a portable/mobile form that is easily cleaned and maintained. For these reasons CXR remains a potentially appealing technology for lung cancer screening programs especially in situations where clinical resources are limited and patients are located far from health infrastructure (WHO, 2021b).

Implementation of the CXR imaging mode along with deep learning techniques to boost radiologist productivity and sensitivity could, in turn, provide a pragmatic and economically feasible mechanism for broad population screening of lung cancer, thereby potentially saving many lives and reducing the economic impact of this disease. A population-wide lung cancer CXR screening program would provide a valuable corpus of training data for lung cancer detection deep learning models, but only if collected images were free of confounding variables and bias with privacy guaranteed. Much has been written in relation to the use of federated learning as a technology to implement privacy by design (Cavoukian, 2012; Wahlstrom et al., 2020), but there has been surprisingly little investigation into the effects of homogenizing medical image data to remove images biases and thereby improve model generalization. This study proposes a novel image pre-processing pipeline that simultaneously homogenizes and debiases chest X-ray images leading to improvements in both internal classification and external generalization, paving the way for federated approaches for effective deep learning lung nodule detection. This is the first study that systematically assesses the utility of combining several debiasing techniques using a process of ablation to determine the impact of each.

## 1.1 Related work

Although there are many studies into the use of deep learning algorithms to classify thoracic disease (Allaouzi & Ben Ahmed, 2019; Bharati et al., 2020; Chen et al., 2019; Dsouza et al., 2019; Ho & Gwak, 2019; Irvin et al., 2019; Ivo et al., 2019; Pham et al., 2020; Wang et al., 2021; Wang et al., 2017), very few of these studies apply any image homogenization techniques apart from the near-universal application of histogram equalization. Lung field segmentation was used by (Liu et al., 2019) to improve both AUC and disease localization for 14 thoracic conditions from the chest X-ray14 dataset (Wang et al., 2017), but no external generalization testing was performed in their study. Lung field segmentation was also used in a study by (Mendoza & Pedrini, 2020) as a pre-processing step in CXR based nodule localization and characterization leading to state-of-the-art results for this task, although ablation testing excluding the segmentation algorithm was not performed. A very small number of studies have considered the effect of rib and bone suppression on automated lung nodule detection (Horváth et al., 2013; Orbán et al., 2010; Simkó et al., 2008). Of these (Gang et al., 2018) is the most recent and comprehensive, being an investigation of deep learning with both lung field segmentation and bone suppression to improve automated nodule detection for the JSRT dataset. This study found a segmented and bone suppressed image corpus led to better deep learning CNN training and validation accuracy when combined with exclusion of outlying records (5% of all images). External validation of these results was not reported.

An alternative line of research into debiasing deep learning CXR automated diagnosis systems is the use of custom loss functions to disentangle features thereby achieving confounder-free training (Robinson et al., 2021; Zhao et al., 2020). Basically, the idea involves including the training data source as a "bias" feature that penalizes gradient descent where that feature allows the model to discriminate the source dataset provenance. In the context of COVID-19 detection (Robinson et al., 2021) it was found that this approach improved AUC by 13% on held-out data, outperforming both histogram equalization and lung field segmentation as debiasing techniques. To date, the most advanced form of a fair confounder free training algorithm is to be found in (Zhao et al., 2020) who allow for a continuously variable confounder values rather than the binary or discrete values typically found in disentanglement training algorithms (Creager et al., 2019; Roy & Boddeti, 2019; Xie et al., 2017). This study successfully applied confounder free learning to several tasks including determining bone age from pediatric hand x-ray images and reducing the effect of sex as a confounding variable. The current state of the art is limited by requisite advanced knowledge of the confounding variable to be suppressed using this technique along with the creation of appropriately conditioned datasets (Zhao et al., 2020).

This study aims to address bias and confounding at the source being the medical images themselves, since in the federated learning context data sources are indeterminate and therefore source-based confounding variables would be too diverse to be mitigated by feature disentanglement approaches alone.

## 1.2 Importance to the field

The recent worldwide COVID-19 pandemic provided the research community with a unique opportunity to validate computer vision-based CXR analysis techniques in a real-world application, since the accessibility of CXR as an imaging mode made it an ideal tool for tracking COVID-19 lung involvement and disease progression. Despite very promising early results, many development studies are overly optimistic because of confounding bias, whereby machine learning-based classifiers chose "shortcuts" over signals especially where disease positive and control sample images are independently sourced (DeGrave et al., 2021). In such cases, source-dependent systematic differences in image attributes such as brightness, contrast, labeling, projection, and patient position can overwhelm the pathological signal differences between the classes, resulting in unreliable classification results for real-world clinical applications. Confounding variables can also come from demographic factors such as sex, socioeconomic status, and age of sample populations (Seyyed-Kalantari et al., 2021). Egregious examples of confounded machine learning-based CXR studies are the numerous COVID-19 diagnostic studies that have reported extremely high diagnostic accuracy for COVID-19 pneumonia against "other" viral pneumonia without recognizing that the COVID-19 sample population consisted of elderly patients, whilst the viral pneumonia control population consisted of pediatric samples, as reported by (Roberts et al., 2021).

The problem of confounding bias is a core inhibitor to the clinical acceptance of machine learning CXR analysis algorithms. The usefulness of such systems is wholly dependent upon the size and quality of labeled training data corpuses upon which they are trained. Systematic differences caused by demographic factors, image acquisition apparatus calibration and operation, projection, and regional morphology diversity will always be present, and it follows that biases in training data corpuses are inherently unavoidable. A practical and reliable approach to CXR homogenization is needed to allow deep learning models to generalize unseen datasets accurately. In this study, we demonstrate an image pre-processing pipeline that homogenizes and debiases chest X-ray images, thereby improving both internal classification results and external generalization, paving the way for low-cost and accessible deep learning-based clinical systems for lung nodule detection.

## 2 Methods

Deep learning models were developed in Python/Tensorflow (Abadi et al., 2016) on the University of Technology Sydney interactive High Performance Computing Environment.

### 2.1 Datasets

A process of bibliographic analysis was undertaken to determine the list of publicly available datasets containing CXR images labelled with lung cancer or conditions associated with lung cancer.

To identify lung cancer-specific datasets, a title/abstract/key search was performed in Scopus using the following query string:

**(TITLE-ABS-KEY(chest AND (x-ray or radiograph) AND (lung or pulmonary) and (nodule or mass or tumor) AND (dataset or database) ))**

This query returned 451 results. A document-based citation analysis was performed using VOSViewer (van Eck & Waltman, 2010) whereby the relatedness of the items was based simply on the number of times items had cited each-other. The reasoning being that the publicly available data sources would be amongst the most highly cited items in this field of research. Of the 451 items, the largest set of connected (by citation) items was 153 resulting in a graph consisting of 7 clusters as shown in Figure 1 below:

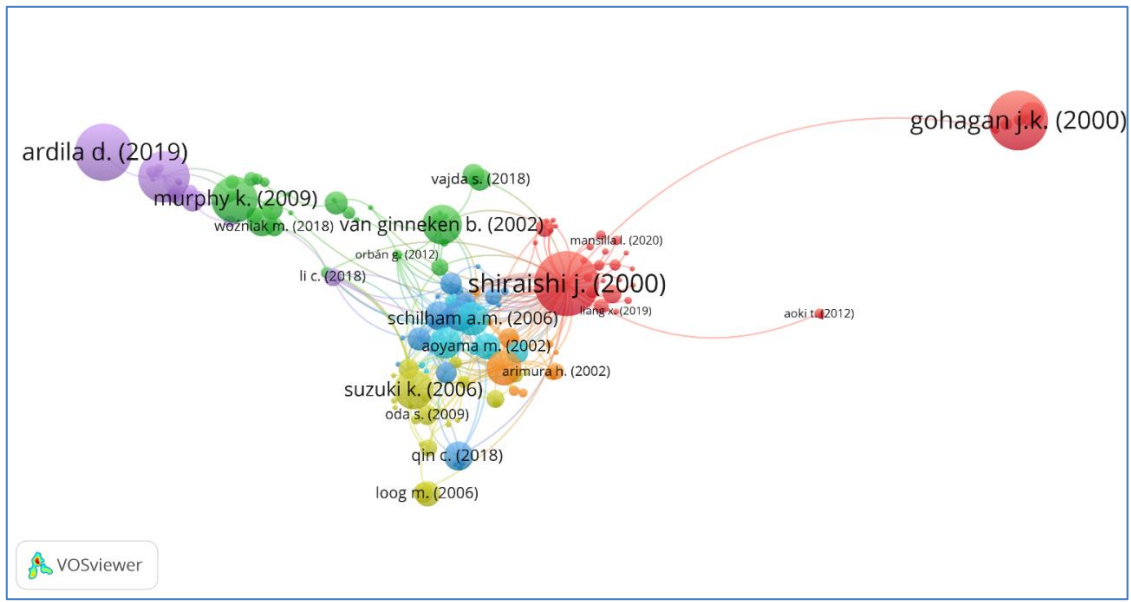

**Fig. 1. Network Visualization for lung nodule dataset query**

Each paper in each cluster was then full text reviewed to determine the underlying dataset used for the study. In addition, common themes within the cluster were identified and logged in table 1.

| Cluster | Colour | Number of papers | Central Paper/s | Imaging Mode | Connecting Theme | Public Data-source/s |
|---|---|---|---|---|---|---|
| 1 | Red | 36 | (Shiraishi et al., 2000),(Gohagan et al., 2000) | CXR | Connected by use of JSRT (Shiraishi et al., 2000) and PLCO (Gohagan et al., 2000) data sets as a lung nodule dataset along with Deep Learning approach for image analysis and nodule classification. | JSRT PLCO |
| 2 | Green | 29 | (Murphy et al., 2009),(Van Ginneken et al., 2002) | CT/CXR | Connected by use local feature analysis, linear filtering, clustering techniques and other non-deep learning techniques. | LIDC |
| 3 | Blue | 24 | (Hardie et al., 2008) | CXR | Artificial Intelligence and machine learning methods including ANN, SVM, KNN. Typically makes use of the JSRT database. | JSRT |
| 4 | Khaki | 23 | (Suzuki et al., 2006) | CXR | Rib/Bone suppression and image enhancement techniques including wavelet transform methods. | JSRT |
| 5 | Purple | 17 | (Ardila et al., 2019),(Setio et al., 2017) | CT | Use of deep learning and shape analysis to diagnose lung cancer from Chest CT images. | Luna16 |
| 6 | Teal | 12 | (Schilham et al., 2006),(Suzuki et al., 2005) | CXR | KNN classification of nodules as blobs. Uses stratification of JSRT to train/calibrate schemes to reduce false positive detection by algorithms. | JSRT |
| 7 | Orange | 12 | (Xu et al., 1997) | CXR | A set of older papers using various techniques to detect nodules and reduce false positive detections | Private Data JSRT |

**Table 1. Description of Published research clusters (lung cancer specific)**

It is clear from this analysis that the most widely studied lung nodule data set is the Japanese Society of Radiological Technology database (JSRT) (Shiraishi et al., 2000) with the CXR subset of Lung Image Database Consortium Image Collection (LIDC) (Armato et al., 2011) along with the PLCO (Gohagan et al., 2000) dataset also used in influential studies. The JSRT dataset is the most cited data source with 447 citations at the time of writing (5 October 2021), and the use of this dataset across numerous studies has resulted in this dataset being the de-facto standard dataset for comparative studies into automated lung nodule detection. For this reason, we have chosen the JSRT dataset as the CXR corpus for training deep learning models for this study. The JSRT dataset also provides gold-standard lung field masks which allow for good quality segmentation of the dataset along with training data for our U-Net (Ronneberger et al., 2015) based lung segmentation algorithm discussed in section 2.2.2. Finally, the JSRT authors conducted an AUC-ROC study using 20 radiologists from 4 institutions determining an average AUC of 0.833 +/- 0.045. This is useful knowledge for calibrating our expectations for model performance as well as sanity checking of our internal testing results. As such, we selected a stable AUC of 80% as the target reference to compare the number of images pruned for successful internal training of each model.

For external testing we are using the CXR subset of the LIDC. This dataset includes 280 CXR images with expert labels indicating the presence and malignancy of lung cancer. This dataset has been labelled by four radiologists with access to corresponding patient CT scans to confirm the lung cancer diagnosis. This dataset is used as an inference-only dataset to check the generalization of the JSRT trained models with various configurations of the pre-processing pipeline.

Additional qualitative and quantitative properties of the JSRT and LIDC datasets are provided in Table 2.

| Data Set | Nodule Image Count | Non-Nodule Image Count | Image Size/Format | Label Accuracy AUC-ROC | Notes |
|---|---|---|---|---|---|
| **JSRT** | 154 images from 154 patients | 93 images from 93 patients | Universal Image Format 2048 x 2048 12-bit grayscale | 20 radiologists from 4 institutions. 0.833 +/- 0.045 | Labelled by 3 x experienced radiologists Includes metadata with 5 levels of subtlety Includes gold-standard lung masks |
| **LIDC** | 280 images from 157 patients | 0 | DICOM Extracted and compressed to 512 x 512 PNG using Pydicom (Mason, 2011) | Not provided | Labeled by 4 radiologists with access to patient CT scan for confirmation |

**Table 2. Summary of data sets used in this study.**

## 2.2 Data pre-processing

### 2.2.1 Histogram equalization

CXR images exhibit variable contrast based on technical calibration of the image acquisition apparatus. Particularly, the energy of the primary beam and the employment of scatter radiation minimization methods such as collimation, grids or air gaps (Murphy & Jones, 2018). Histogram equalization processing has proven to be effective in normalizing the gray-level distribution of CXR images, resulting in uniform image histograms (Shuyue et al., 2006). Histogram equalization has been applied to all images in this study as a standard first step towards image homogenization. This is required since computer vision for deep learning algorithms are trained on continuous pixel intensity values, and systematic differences in image contrast between the datasets would result in biased model training and negatively impact model generalization.

The before and after examples of image histogram equalization for the LIDC and JSRT datasets are shown in figure 1. Prior to histogram equalization, these datasets are easily distinguished from each other due to the overwhelming difference in the brightness and contrast of the images (figure 1 (a) and (c)). As such, the JSRT dataset is evidently overexposed with lower contrast/higher brightness as compared to the LIDC dataset. The application of histogram equalization effectively removes this systematic brightness and contrast differences (figure 1 (b) and (d)).

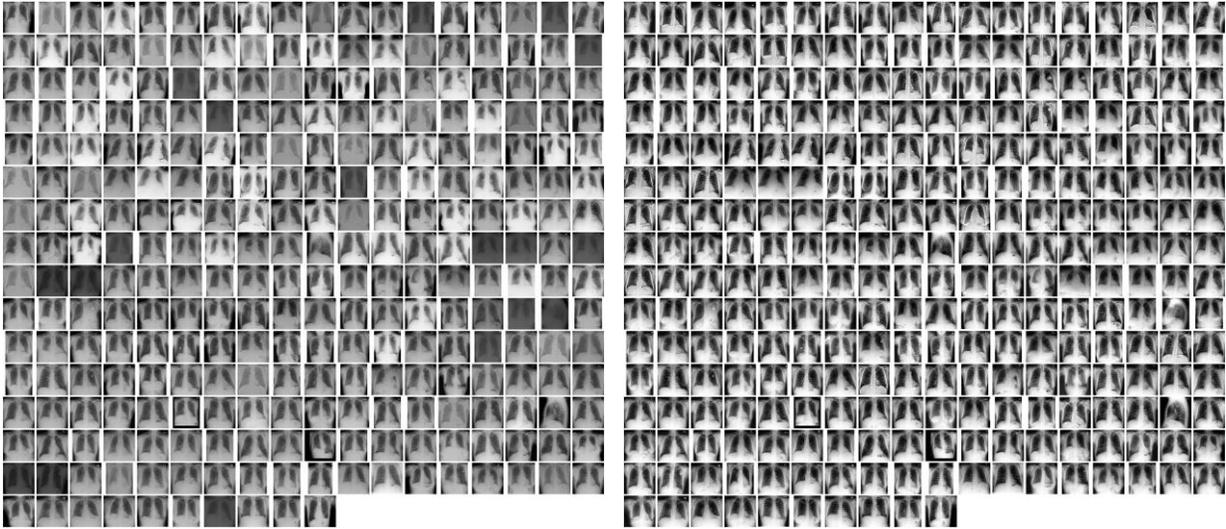

(a) LIDC dataset prior to histogram equalization      (b) LIDC dataset after histogram equalization

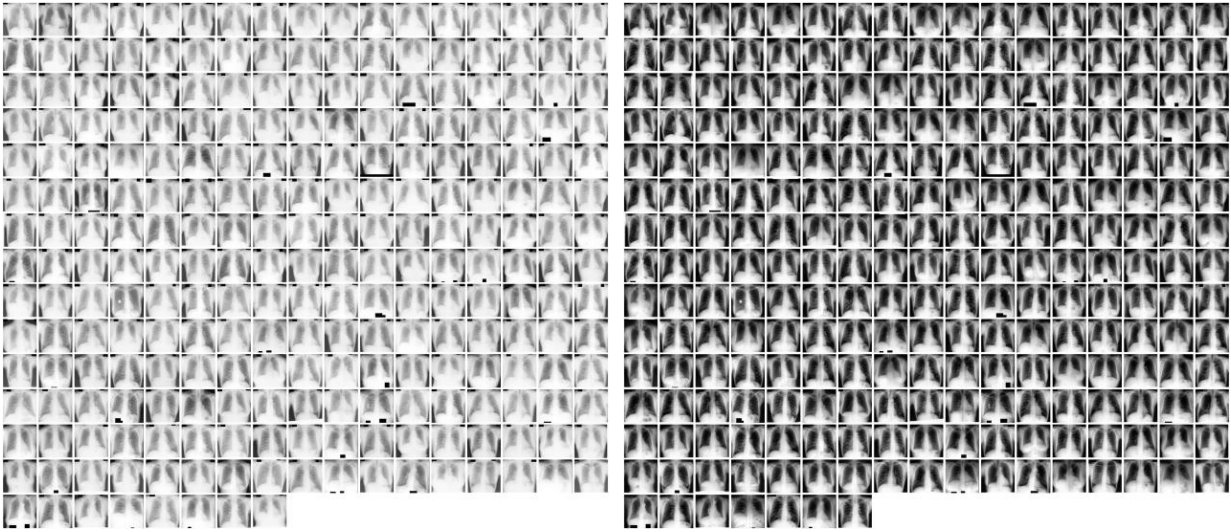

(c) JSRT dataset prior to histogram equalization        (d) JSRT dataset after histogram equalization

**Fig. 2. LIDC dataset with (a) and without (b) image histogram equalization applied.**

### 2.2.2  Lung field segmentation

Lung nodules occur within the lung field therefore non-lung field pixels are signal noise. CXR features such as projection labels and portability markers at the edges of CXR images have been shown by several studies to be a source of confounding bias in deep learning-based medical imaging applications (Badgeley et al., 2019; DeGrave et al., 2021; Zech et al., 2018). Projection labels have been shown to be effective predictors of CXR repository sources with an accuracy of up to 99.98%, and 100% accuracy in separating emergency room vs inpatient CXR images (Zech et al., 2018). The use of bedside mobile CXR apparatus is frequently associated with more severe disease (due to lack of patient mobility) allowing deep learning models to use apparatus portability labels as a signal indicating the disease and its severity. Thus, the elimination of non-lung field pixels using segmentation effectively removes training bias and results in more generalizable deep learning models.

The utility of this approach in medical image classification is supported in previous studies (Rabinovich, 2020; Teixeira et al., 2021). A comprehensive review of lung area segmentation techniques may be found in (Candemir & Antani, 2019). The best performance reported by this survey was achieved using a CNN-based deep learning system with a dice similarity coefficient of 0.980 (Hwang & Park, 2017). A much simpler approach leveraging a U-Net architecture (Ronneberger et al., 2015) was trained on the JSRT dataset (Shiraishi et al., 2000) consisting of 385 CXR images with gold-standard masks to achieve a dice similarity co-efficient of 0.974 (Novikov et al., 2017). We reasoned that the U-Net based architecture could be improved in terms of dice similarity co-efficient as well as adaptability using training data from two additional data sources, the Montgomery and Shenzen datasets (Jaeger et al., 2014) thereby creating a combined training data corpus of 1185 CXR image/mask pairs.

After experimenting with VGG-like CNN (Simonyan & Zisserman, 2015), DenseNet (Huang et al., 2017) and ResNet (He et al., 2015) based U-Net architectures, we found that a Deep Residual U-Net following (Zhang et al., 2018) provided the highest dice similarity scores and is consistent with other studies (Alom et al., 2019; Zuo et al., 2021) outside the medical imaging domain. Since nodules are typically small, we expanded the image size for our residual U-Net implementation from 224 x 224 pixels as used in the original study (Zhang et al., 2018) to 512 x 512 pixels by addition of extra encoder/decoder residual blocks to maximize the image size with reasonable processing time on our computing environment. This allows the segmented image to retain important anatomical features for downstream classification.

This combination of training data sets and network design achieved a maximum validation dice similarity co-efficient of 0.988 at epoch 93 which is a small improvement from (Munawar et al., 2020) who reported a result of 0.974 for lung field segmentation using JSRT, Shenzen, and Montgomery datasets, and (Novikov et al., 2017) who used the JSRT dataset only, as sources. Training loss and accuracy curves along with Dice charts for our residual U-Net are included in figure 3.

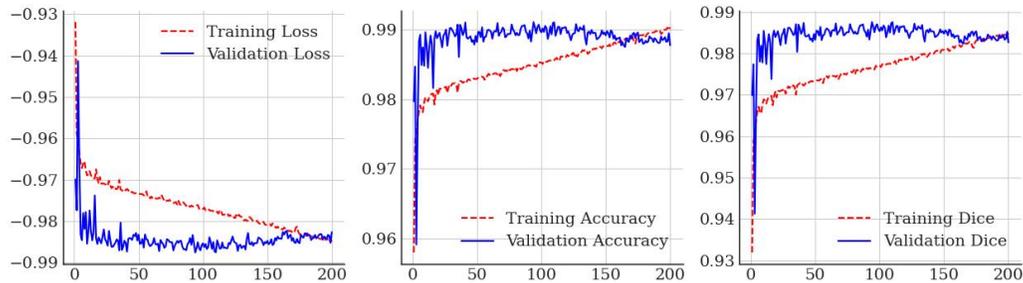

**Fig. 3. Training curves for deep residual U-Net lung field segmentation.**

When applied to the unseen LIDC dataset, the segmentation network did produce minimal artefacts similar to those reported in (Souza et al., 2019) in 6 of the 280 source images as shown in figure 4. These artefacts take the form of pinholes in the generated lung mask. These pinholes were eliminated using morphological closing with a kernel size of 8 followed by flood fill of any contour smaller than a parameterized minimum square area which was determined by experiment to be optimal at 1/16$^{th}$ of the image area. The before and after results of the parameterized morphological closing/contour filling operations are shown in figure 4.

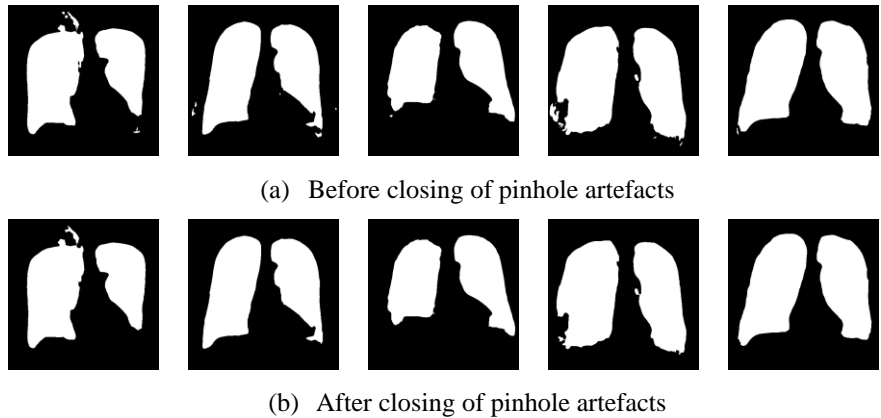

(a) Before closing of pinhole artefacts

(b) After closing of pinhole artefacts

**Fig. 4. Results of morphological closing and parameterized contour filling to complete lung masks for LIDC images**

A montage of LIDC lung field segmentation masks is shown in figure 5. Four image masks (highlighted red) were considered failures for either having a contour count less than 2, or poor visibility of a single lung field. The images associated with these masks were not segmented. Instead, they were removed from the LIDC dataset for generalization testing experiments.

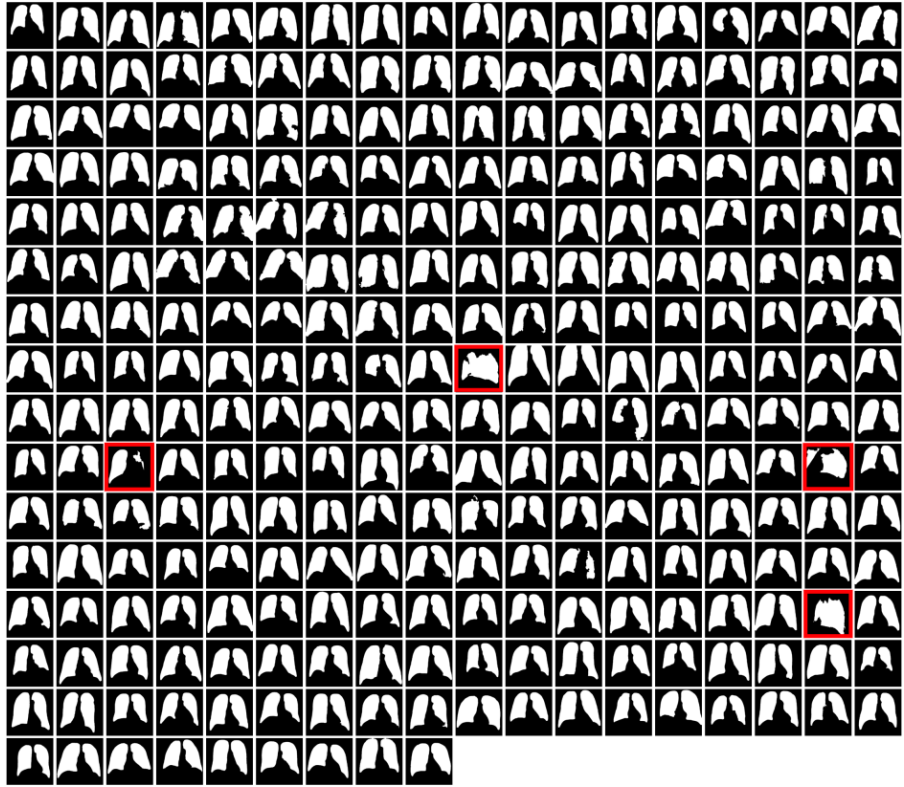

**Fig. 5. LIDC Lung Field Masks automatically generated using hybrid U-Net and single parameter morphological algorithm.**

### 2.2.3 Segmented Lung Field Cropping

Following the segmentation of the CXR images using gold standard masks for JSRT and U-Net generated masks for LIDC, the size of the lung field in the images was observed to be highly variable due to the differences in patient morphology and main beam focal distance as evident from figure 6(a). Note that the area surrounding the lung field contains no useful information and could potentially be a source of confounding bias. Therefore, the CXR images were close cropped on all sides to ensure that the first non-back pixel is on the border of the image for all directions per figure 6(b).

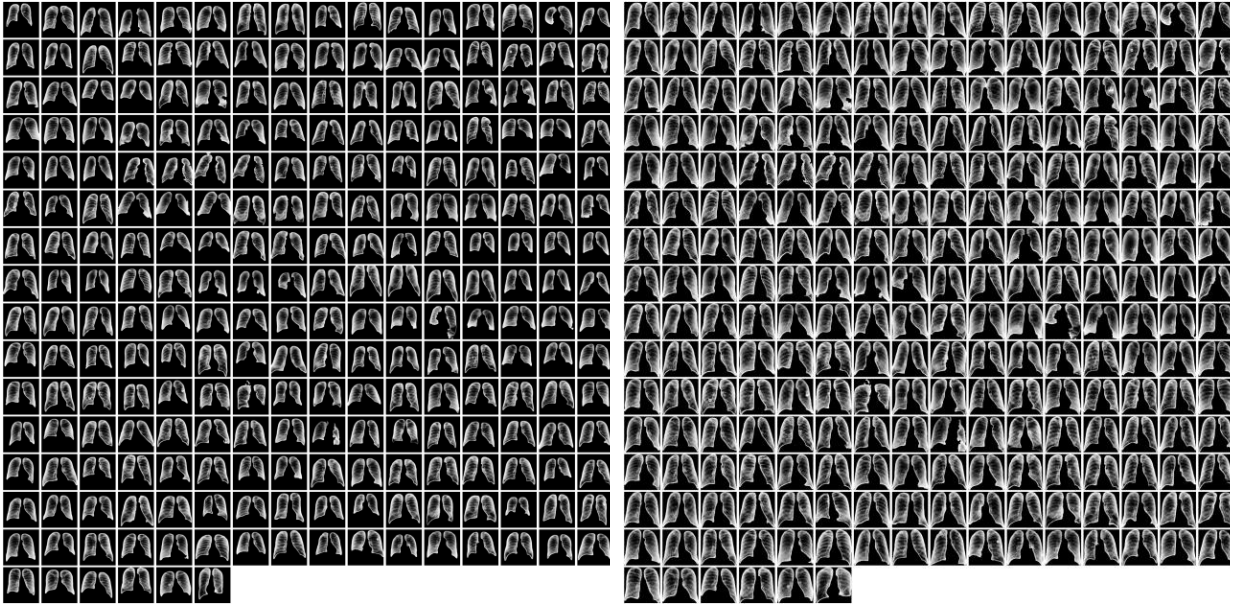

(a) Segmented JSRT images  (b) Segmented and cropped JSRT images

**Fig. 6. JSRT Lung field masks before cropping (a) and after cropping (b)**

### 2.2.4 Rib and Bone Suppression

Following equalization and segmentation we noticed that the ribcage features are very pronounced on the CXR images. Concerningly, ribs on the LIDC dataset generally appeared to be more prominent than ribs in the JSRT dataset. Therefore, we considered the appearance of ribs as a potential confounding variable and sought to suppress and homogenize the appearance of ribs on the CXR images. Two approaches were attempted, firstly, an autoencoder based suppression algorithm is trained using the bone suppressed JSRT images provided by (*BME-MIT-OMR JSRT Database*, 2021). Secondly, we adopt a simple CNN based approach following (Gusarev et al., 2017) by reusing code provided to the community by (*Deep-Learning-Models-for-bone-suppression-in-chest-radiographs*, 2021). As a result we found that the second approach resulted in much sharper images with less blurring on the fine lung structure details which is consistent with the results of (Gusarev et al., 2017). The before and after rib suppressed examples for the raw CXR images are presented in figure 7 (a-d).

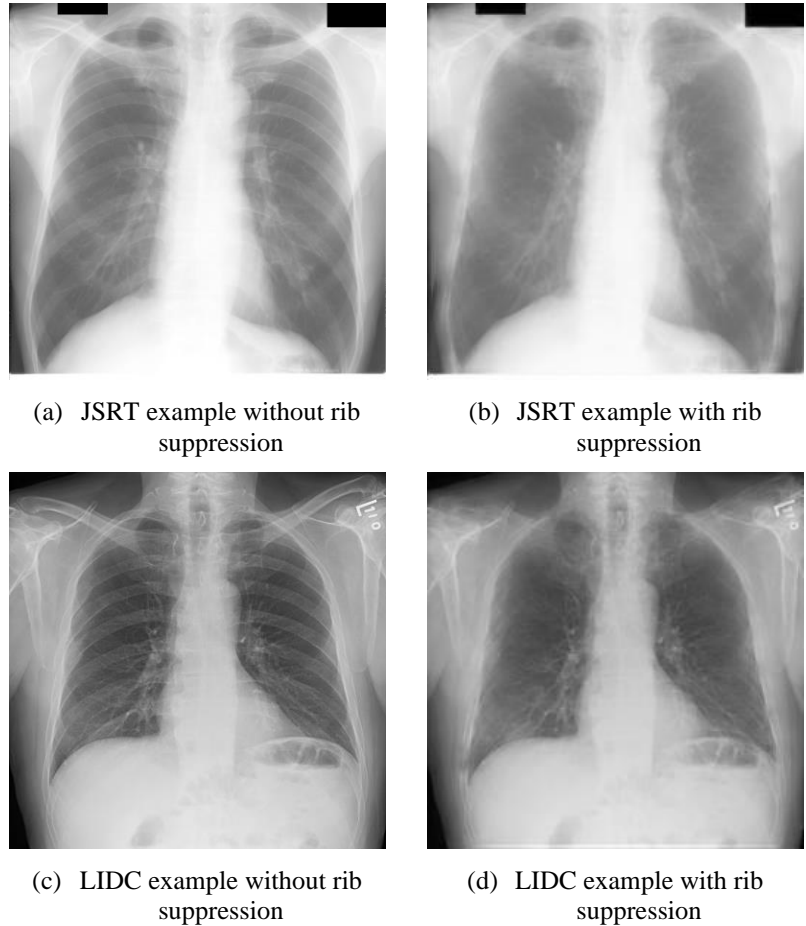

| (a) JSRT example without rib suppression | (b) JSRT example with rib suppression |
| (c) LIDC example without rib suppression | (d) LIDC example with rib suppression |

**Fig. 7. Effect of Rib Suppression for JSRT and LIDC datasets. Fine vascular detail has been preserved well with minimal overall reduction of image sharpness.**

### 2.2.5 Deep Learning Model

The JSRT image dataset was loaded into data/label arrays and randomly split into 4 folds using the scikit-learn StratifiedKFold function (Pedregosa et al., 2011). A VGG16 (Simonyan & Zisserman, 2015) classifier initialized with ImageNet (Deng et al., 2009) weights was selected following experimentation with VGG16 (Simonyan & Zisserman, 2015), VGG19 (Simonyan & Zisserman, 2015), DenseNet-121 (Huang et al., 2017), and ResNet50 (He et al., 2015). Of these candidate classifiers the pre-trained VGG16 model showed the least tendency to overfit and provided the most uniform training results for each fold for the JSRT dataset. This is consistent with our earlier results on small medical image datasets where there is insufficient data to train deeper networks (Horry et al., 2020).

The pre-trained VGG16 classifier was modified with a hand-crafted output head designed to regularize the network and facilitate binary classification of nodule and non-nodule image classes. This output head is shown in figure 8.

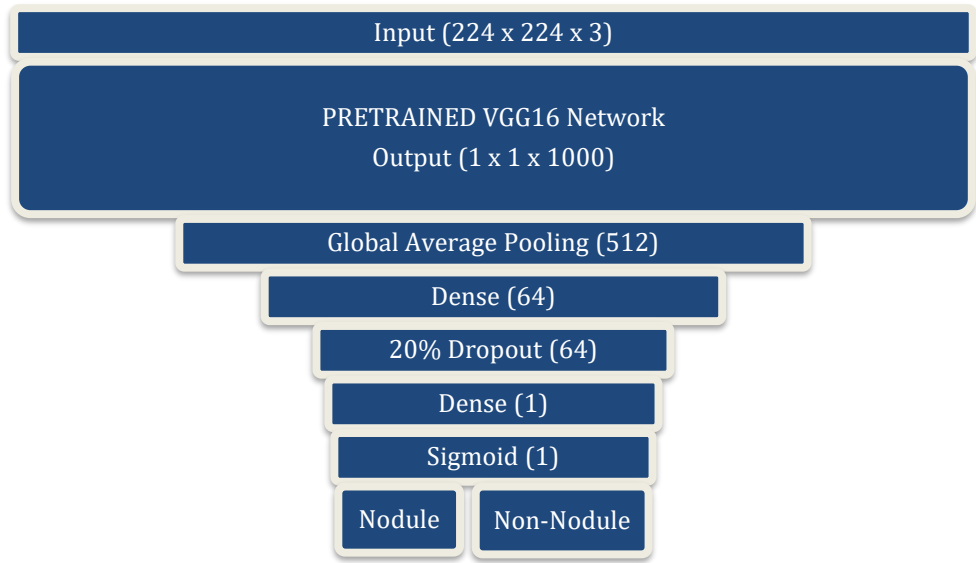

**Fig. 8. Pre-trained VGG16 Classifier with modified output head using sigmoid activation for binary classification.**

The modified VGG16 classifier was trained for each of the four folds using a binary cross entropy loss function with a learning rate of 0.0005 for 50 epochs followed by 10 fine-tuning epochs at a lower learning rate of 0.00001 to avoid overfitting. Trained models were captured at the point of minimum validation loss. Sample training and fine-tuning accuracy/loss curves are shown in figure 9.

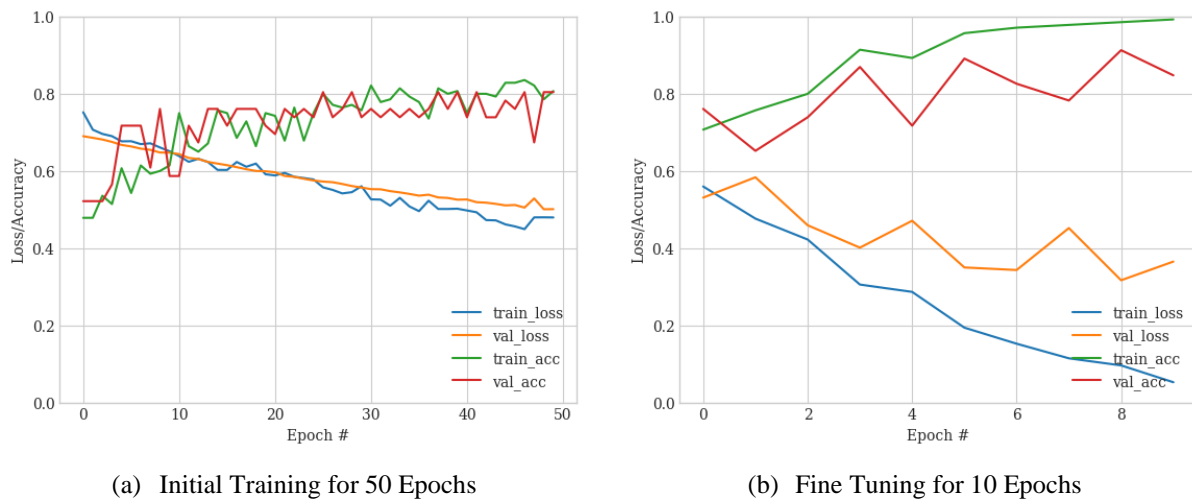

(a) Initial Training for 50 Epochs        (b) Fine Tuning for 10 Epochs

**Fig. 9. VGG16 Training/Loss Curves for training and fine-tuning.**

### 2.2.6 Evolutionary pruning algorithm

During initial 4-fold training/validation we observed great variation in the validation accuracy of the per-fold models despite model and training hyperparameters being held constant. This variability is owing to the randomized images per fold producing some dataset shards being trained more successfully than others.

To identify nodule image samples that were least informative for the nodule class an inference operation was performed against each Nodule CXR image in the JSRT dataset following the completion of each training round. This inference operation consisted of testing all four shards against the k-fold trained model, resulting in 16 inferences per nodule positive image (4 data shards x 4 models). For each round, the JSRT Nodule image with the highest number of misclassifications was added to a prune list and ignored on the next training round, thereby under-sampling the majority Nodule image class. This process was repeated for a total of 61 times, and the number of nodule positive images was reduced from 154 to 93 to balance the non-nodule image sample number. The pruned JSRT dataset was then used to train a master model for external generalization testing against the LIDC dataset. This end-to-end process of pruning and external testing is illustrated in figure 10.

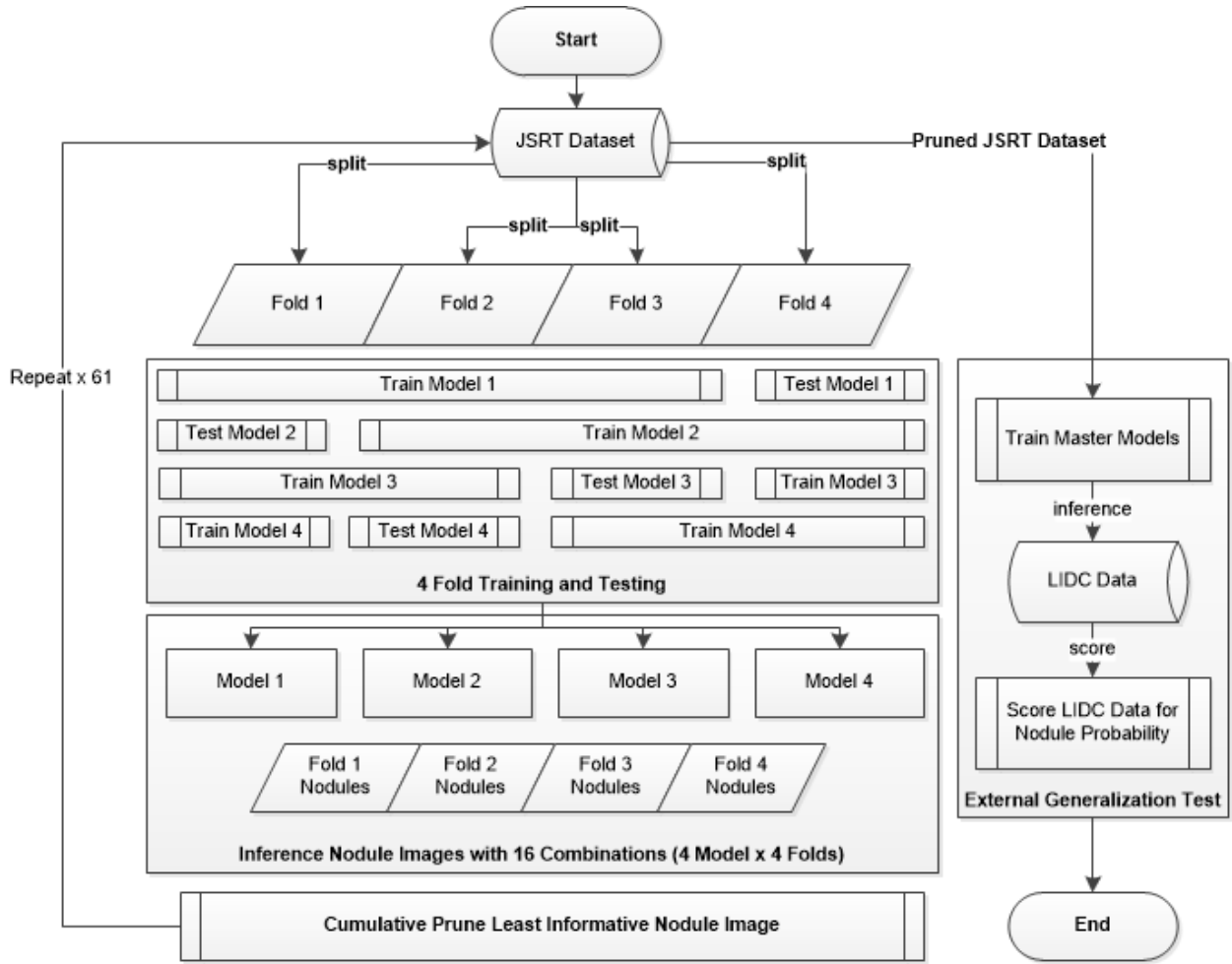

**Fig. 10. 4-fold training/pruning and external testing flowchart.**

### 2.2.7   Experiment setup and ablation studies

Excluding image histogram equalization which is a standard image processing technique that was implemented in line with data ingestion, this study was concerned with any potential effects from the mix and match combinations of lung field segmentation, close cropping, and rib suppression on internal and external classification results.

This resulted in six experiment configurations as detailed in table 3. Note that the cropping operation is only applied in combination with the segmentation operation (Experiment E and F). Experiments A, C, and E did not perform rib suppression, while experiments B, D, and F employed rib suppression. Experiments C and D applied lung field segmentation as a proposed debiasing process, and experiments E and F further apply cropping to the segmented lung field. Each experiment may be considered an ablation study of the complete debiasing pipeline represented by experiment F.

| Experiment | Segmentation | Cropping | Rib Suppression | Sample Image (JSRT JPCLN008) |
|---|---|---|---|---|
| A | False | False | False | 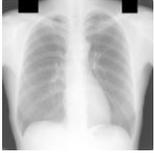 |
| B | False | False | True | 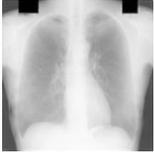 |
| C | True | False | False | 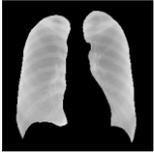 |
| D | True | False | True | 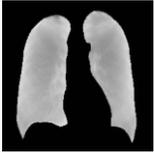 |
| E | True | True | False | 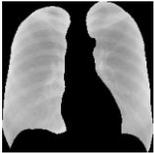 |
| F | True | True | True | 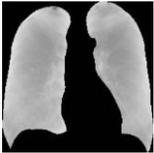 |

Table 3. Experiment/Ablation Tests covering each element of the proposed debiasing pipeline.

## 3   Results

### 3.1  Internal 4-Fold Training and Testing

#### 3.1.1   Internal testing result A (no debiasing operations)

Internal testing results for experiment A which represents training with raw data are presented in figure 11. The initial training round achieved an AUC of around 67.6% +/- 4.1%, which is consistent with (Wang et al., 2017) and another deep learning study on the JSRT dataset (Ausawalaithong et al., 2019). These results improve significantly as the poorest performing nodule test samples are pruned, achieving a stable 80% at 15 pruned records and 93.5%+/-2.4 when the classes are balanced with 62 pruned records.

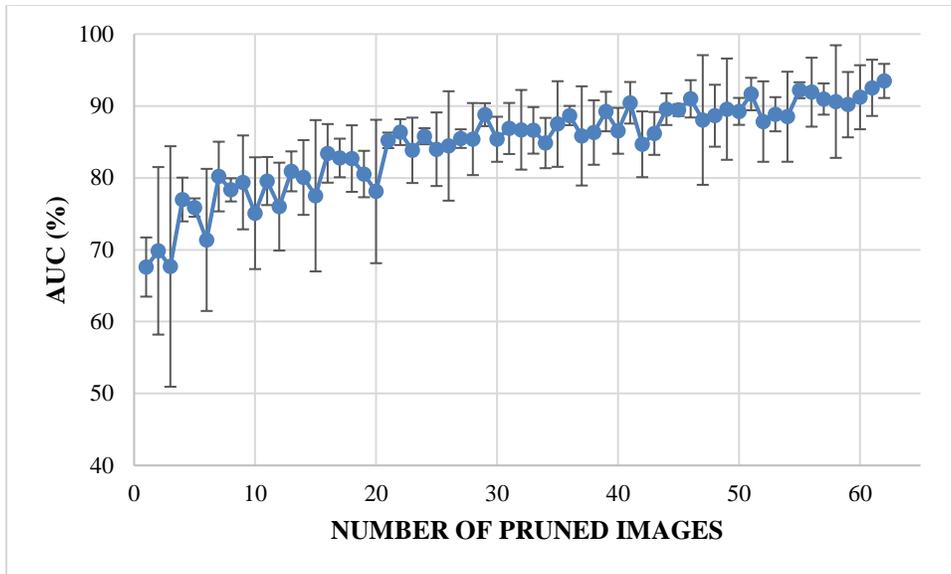

**Fig. 11. Internal testing results on histogram equalized images (experiment A).**

### 3.1.2 Internal testing result B (rib suppression operator)

Internal testing results following the application of the rib suppression operator only are shown in figure 12. Rib suppression resulted in a significant improvement of initial results to 70.4 +/- 5.0%. Once again, results improve linearly as the poorest test nodule samples are pruned, with stable 80% AUC achieved at around 15 pruned records with an AUC of 92.6+/-4.6% achieved at the point of class balance.

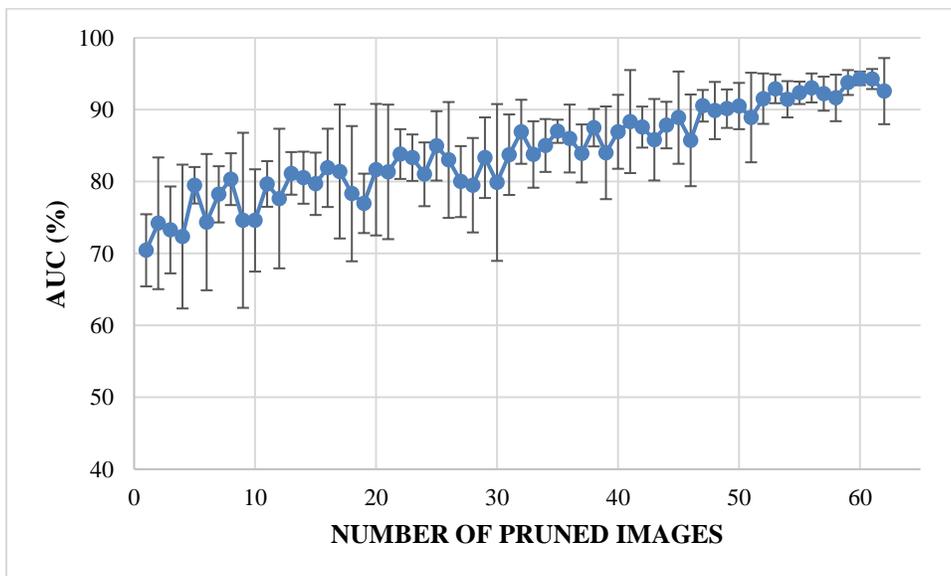

**Fig. 12. Results on histogram equalized and rib suppressed images (experiment B).**

### 3.1.3 Internal testing result C (segmentation operator)

Application of the lung field segmentation operator resulted in a further improvement to initial internal testing results with an AUC of 72.6 +/- 1.6 % achieved without record pruning. Results improved following pruning of the poorest performing lung nodule test images, however, at a lower rate than both experiments A and B as stable AUC of 80% is not reached until around 35 pruned records. This slower rate of AUC improvement per pruned image can be interpreted as improved classification sensitivity for the more challenging nodule images. Despite the improved initial AUC, the slower rate of improvement results in a lower AUC of 88.1+/-1.0 at the point of class balance. The lower final AUC along with a very low error margin may be caused by the removal of confounding variables outside the lung field that was allowing models A and B to take shortcuts in learning.

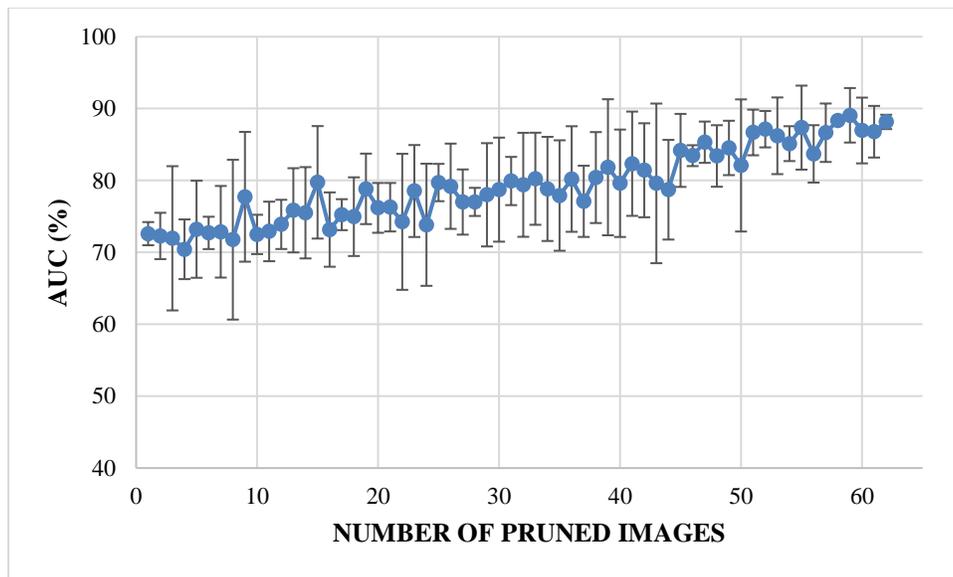

**Fig. 13. Results on histogram equalized and lung field segmented images (experiment C).**

### 3.1.4 Internal testing result D (segmentation + rib suppression operators)

Application of both lung field segmentation and rib suppression to the histogram equalized images resulted in another incremental improvement in the initial testing round with AUC of 74.9+/-3.9% with no pruned records as shown in figure 14. A stable AUC of 80% was reached after pruning only 13 records representing a significant improvement over rib suppression alone (experiment C). AUC of 90.3 +/-4.1% was achieved at the point of class balance, again a significant improvement on segmentation alone.

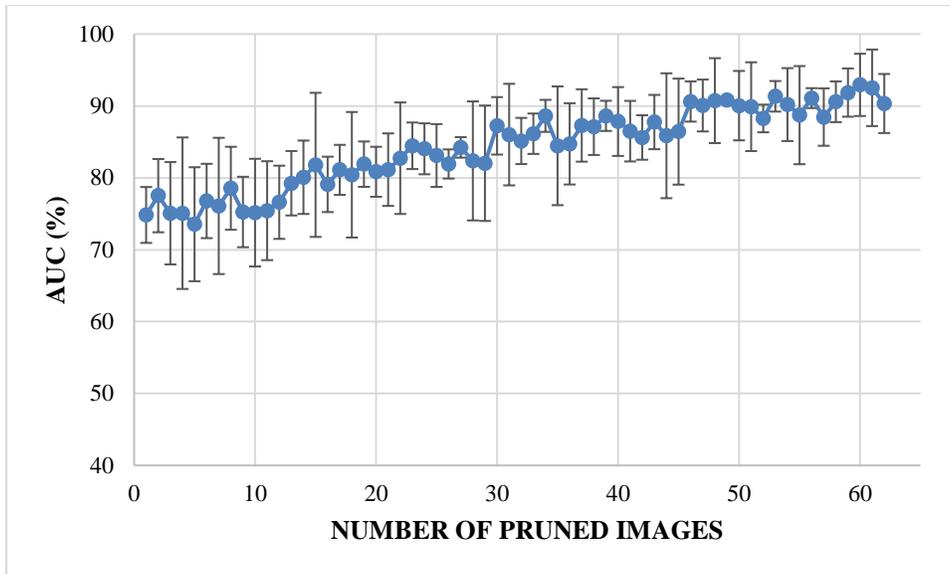

**Fig. 14. Results on histogram equalized, rib suppressed and lung field segmented images (experiment D).**

### 3.1.5 Internal testing result E (segmentation + cropping operators)

Compared to segmentation only, the cropping operation (figure 15) reduced the initial AUC to 70.7 +/- 11.4% but allows the model to reach stable 80% AUC at around 25 pruned records (compared to 35 for segmentation only). AUC of 91.0+/-2.4 was achieved at the point of class balance with 62 pruned records which is much higher than the final AUC of 88.1+/-1.0 achieved using segmentation alone.

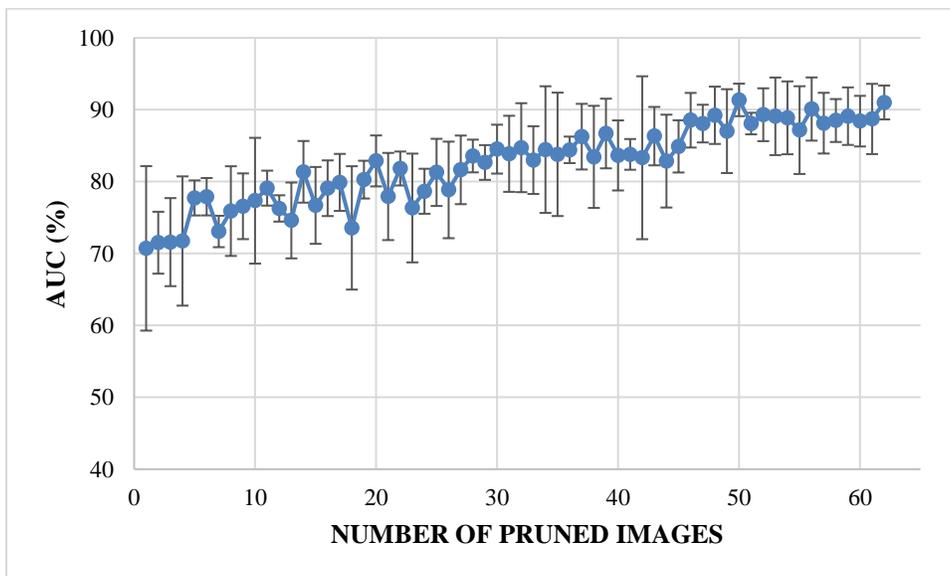

**Fig. 15. Results on histogram equalized and cropped lung field segmented images (experiment E).**

### 3.1.6 Internal testing result F (segmentation + cropping + suppression operators)

The best result was achieved in experiment F using all operators per figure 16. Initial average AUC was 74.2 +/- 7.1%. This experiment however achieved a stable 80% AUC at 10 pruned records with an AUC of 90.5+/-4% achieved at the class balance point of 62 pruned nodule images. Although other experiments may have achieved higher AUC values, this was only possible after discarding far more difficult nodule images. Experiment F provided the best balance of excellent initial results, with minimal pruning of only 10 of 293 or 3.4% of nodule images to reach the target stable 80% AUC.

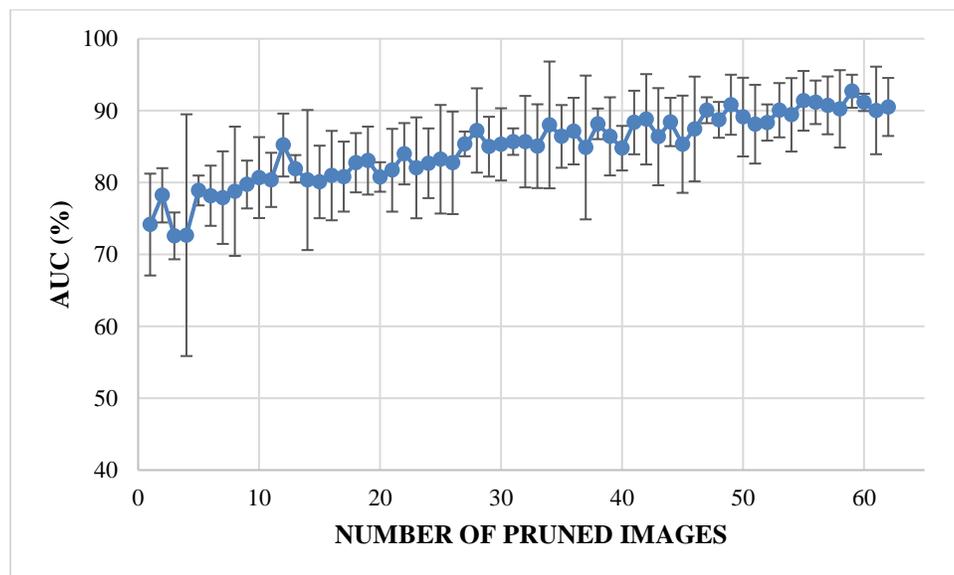

**Fig. 16. Results on histogram equalized, rib suppressed and cropped lung field segmented images (experiment F).**

### 3.1.7 Pruned Records Analysis (from experiment F)

Since experiment F achieved excellent results with minimal pruned nodule records, we were interested in the attributes of the pruned records potentially leading our trained CNN to misclassify the nodule images as non-nodule. We consulted a practicing radiology registrar for the interpretation of the top five misclassified JSRT nodule images as logged in table 4. Each of the images that were found to be difficult by evolutionary pruning was also challenging for a human reader by virtue of visibility threshold or nodule features hidden behind the cardiac silhouette and/or position overlapping bone, vascular marking, or breast tissue.

| Filename | Image | JSRT Metadata Notes | Radiology Registrar Observations |
|---|---|---|---|
| **JPCLN151.png** | 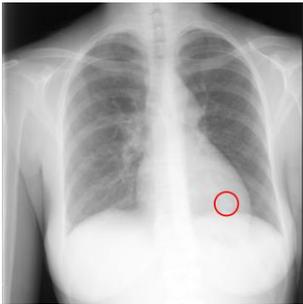 | Extremely Subtle<br>14 mm<br>(1520,1364). L lower. | Extremely subtle<br>Behind cardiac silhouette<br>Overlaps vascular marking |
| **JPCLN003.png** | 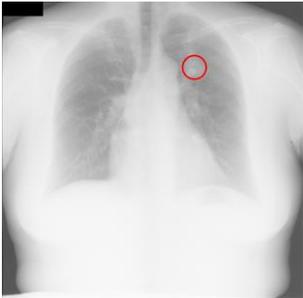 | Obvious<br>30 mm<br>(1303,447). L upper. | Obvious<br>Overlaps vascular markings |
| **JPCLN130.png** | 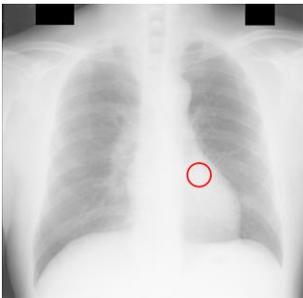 | Extremely Subtle<br>30 mm<br>(1335,1163). L lower. | Extremely subtle<br>Behind cardiac silhouette |
| **JPCLN141.png** | 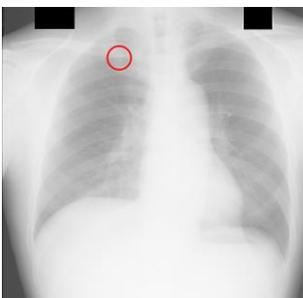 | Extremely Subtle<br>10 mm<br>(794,346). R. upper. | Extremely subtle<br>Behind rib/clavicle |
| **JPCLN142.png** | 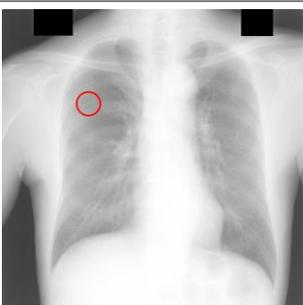 | Extremely Subtle<br>10 mm<br>(586,643). R. upper. | Not visible. |

**Table 4. Top 5 most difficult JSRT Nodule images pruned for experiment E. Pruning these images results in stable AUC of around 78%. Location of Nodule per JSRT metadata is represented by a circle. All nodules are in challenging positions even for human radiologists except for JPCLN142.png which is at the threshold of human visibility.**

### 3.2 External testing evaluation using pruned models

A series of master models were trained for each experiment using JSRT based datasets that had been balanced by pruning the over-represented nodule classes, keeping only the images that were most informative of the nodule classification by testing over 4-folds as described above. This resulted in 6 fully trained models which were then used to inference Nodule scores for each image in the LIDC CXR subset. All CXR images in the LIDC dataset are lung cancer cases, with nodules identified by CT scan and sometimes very subtle or invisible to human readers on the CXR image. This final phase of the experiment was conducted to confirm which image processing operators had promoted the generalization of the JSRT trained models to an independent dataset. Results are presented in figure 17 where the nodule class probability has been plotted for each LIDC image in a scatter-plot. Since all LIDC images are of lung cancer patients, the plot can be expected to have the majority of datapoints above 0.5 probability if the JSRT model has successfully generalized to the LIDC dataset. This is the case for models implementing the rib-suppression operator (experiments B, D, and F), particularly where the lung field segmentation operator is also used (experiments D and F) with best results achieved in experiment F using the combination of rib suppression, lung field segmentation and close cropping operators. This combination of operators achieved a classification accuracy of 89% in LIDC nodule images which we consider to be an excellent result given the subtlety of nodule features in some of the LIDC images; some nodules were measured to be less than 3 mm on CT scan (Armato et al., 2011), and smaller nodules were poorly visible on the corresponding CXR.

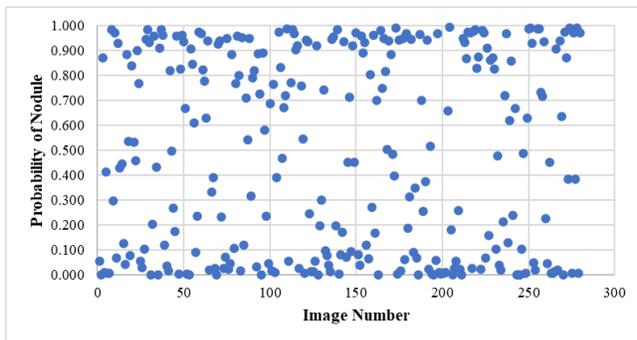
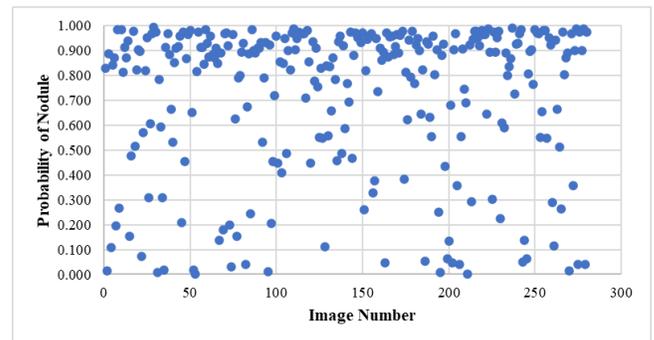

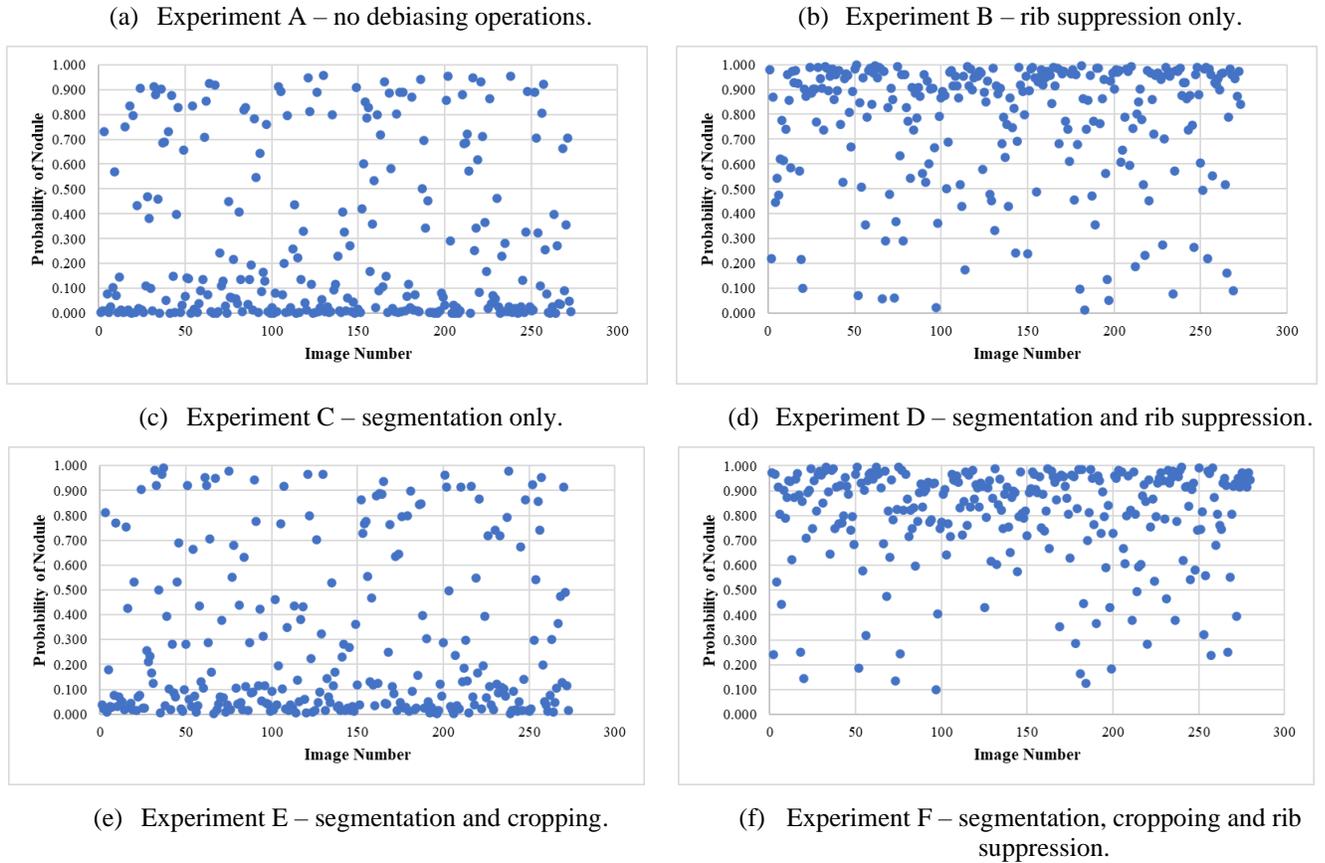

(a) Experiment A – no debiasing operations.  (b) Experiment B – rib suppression only.

(c) Experiment C – segmentation only.  (d) Experiment D – segmentation and rib suppression.

(e) Experiment E – segmentation and cropping.  (f) Experiment F – segmentation, croppoing and rib suppression.

Fig. 17. Results of external testing against the LIDC dataset using JSRT trained models with all combinations of proposed debiasing operators showing effectiveness of rib suppression in improving classification results (b), (d), and (e) especially when lung field segmentation is also applied (d) and (e).

## 4 Discussion

These experiments delivered promising and surprising results in equal measure. The first key observation is that use of image histogram equalization alone, with no further debasing operators resulted in models that did not generalize at all per figure 17(a). This is of concern given that most studies implementing deep learning in automated analysis of CXR images use histogram equalization as the only pre-processing step. The second interesting result was that lung field segmentation alone actually worsened the classification metrics both internally and externally. We see this as strong evidence of confounding variables present in the JSRT image data corpus that allowed some degree of shortcut learning from unsegmented images, which was eliminated when these images were segmented. The poor external testing result provides evidence that lung field segmentation alone is not enough to sufficiently debias an image corpus to promote generalization.

Our best generalization results were achieved using the rib suppression operator. Although this technique has been shown to improve human radiologist sensitivity to nodules (von Berg et al., 2016), there is little published literature that is interested in the use of rib and bone suppression to promote deep learning model generalization for medical images. To the best of our knowledge, this is the first study to do so. Of particular interest was that lung field segmentation with and without close cropping improved external classification results, but only after the application of the rib suppression operator. Our explanation for this behavior is that the appearance of ribs in CXR images is a confounding variable itself, which matched our observations that the visibility of ribs in CXR images varies greatly between datasets. As such, improvement in generalization only becomes evident after lung field segmentation and cropping when this confounder is suppressed.

### 4.1 Limitations

This study was limited by datasets made publicly available by JSRT and LIDC. Given that access to CXR data is governed by privacy concerns, the future progress of this study will be undertaken using a federated learning paradigm in conjunction with a clinical study. This will provide real-world validation of the proposed debiasing pre-processing pipeline in preparation for the development of an automated lung nodule detection reference implementation.

# 5 Conclusion

The promise of private, federated deep learning to improve access to training data whilst preserving privacy can only be realized in the presence of an effective debiasing image pre-processing pipeline. We have shown that the combination of histogram equalization, rib suppression, and close-cropped lung field segmentation effectively homogenizes and debiases a corpus of CXR images allowing trained models to generalize to external datasets utilizing the same image pre-processing pipeline. This is the first study to include a rib suppression operator in an external generalization study, with this operator being found to be essential in achieving external generalization accuracy of 89% for a model trained on the JSRT dataset and tested against the external LIDC dataset. This state-of-the-art result provides the community with confidence that deep learning classifiers can be trained in a bias-free manner to apply across datasets of different providence thereby paving the way for useful clinical tooling with broad application. We foresee that such a tool could be used to achieve a lower cost/risk lung cancer screening using the CXR imaging mode, resulting in better access to earlier diagnosis, potentially saving many lives and reducing the societal costs of the most deadly cancer.

**Funding**

No funding has been received for the development or documentation of this work.

**Declaration of competing interest**

The authors declare that they have no competing interests in association with this work.

**Acknowledgements**

The first author acknowledges their employer, IBM Australia Pty Limited, for providing flexibility allowing for the progression of experimental work and documentation of results. Computational facilities were provided by the UTS eResearch High Performance Compute Facilities.